\documentclass{article}

\usepackage{arxiv}

\usepackage[utf8]{inputenc} % allow utf-8 input
\usepackage[T1]{fontenc}    % use 8-bit T1 fonts
\usepackage{hyperref}       % hyperlinks
\usepackage{url}            % simple URL typesetting
\usepackage{booktabs}       % professional-quality tables
\usepackage{amsfonts}       % blackboard math symbols
\usepackage{nicefrac}       % compact symbols for 1/2, etc.
\usepackage{microtype}      % microtypography
\usepackage{lipsum}
\usepackage{graphicx}

\usepackage{algorithm,algorithmicx,algpseudocode}
\usepackage{amsmath,amssymb}
\usepackage{booktabs,multirow}
\usepackage{siunitx}
\usepackage{subcaption}
\usepackage{makecell}
\usepackage{pifont}

\graphicspath{ {./images/} }

\title{Universal Adversarial Suffixes Using Calibrated Gumbel–Softmax Relaxation}

\author{
Sampriti~Soor \\
Center for Intelligent Cyber Physical Systems\\
Indian Institute of Technology Guwahati, India\\
\texttt{sampreetiworkid@gmail.com} \\
\And
Suklav~Ghosh \\
Department of Computer Science and Engineering\\
Indian Institute of Technology Guwahati, India\\
\texttt{suklav@iitg.ac.in} \\
\And
Arijit~Sur \\
Department of Computer Science and Engineering\\
Indian Institute of Technology Guwahati, India\\
\texttt{arijit@iitg.ac.in} \\
}

\begin{document}
\maketitle
\begin{abstract}
Language models (LMs) are often used as zero-shot or few-shot classifiers by scoring label words, but they remain fragile to adversarial prompts. Prior work typically optimizes task- or model-specific triggers, making results difficult to compare and limiting transferability. We study universal adversarial suffixes: short token sequences (4–10 tokens) that, when appended to any input, broadly reduce accuracy across tasks and models. Our approach learns the suffix in a differentiable “soft” form using Gumbel--Softmax relaxation and then discretizes it for inference. Training maximizes calibrated cross-entropy on the label region while masking gold tokens to prevent trivial leakage, with entropy regularization to avoid collapse. A single suffix trained on one model transfers effectively to others, consistently lowering both accuracy and calibrated confidence. Experiments on sentiment analysis, natural language inference, paraphrase detection, commonsense QA, and physical reasoning with Qwen2-1.5B, Phi-1.5, and TinyLlama-1.1B demonstrate consistent attack effectiveness and transfer across tasks and model families.
\end{abstract}

% keywords can be removed

%%%%%%%%%
%%%%%%%%%
%%%%%%%%%
\section{Introduction}
\label{sec:intro}
Language models (LMs) are increasingly deployed in safety-critical and user-facing applications, where even subtle vulnerabilities can have significant consequences. While these models demonstrate remarkable fluency and reasoning ability \cite{radford2019language, brown2020language}, their predictions are often unstable under small input modifications. This fragility poses not only a reliability concern for downstream tasks such as sentiment analysis, natural language inference, and commonsense reasoning, but also a security risk when adversarial actors deliberately attempt to manipulate model outputs \cite{biggio2018wild,ren2019generating}. A growing body of research highlights that language models can be misled by crafted perturbations that remain imperceptible or irrelevant to humans \cite{morris2020textattack,zou2023universal}, emphasizing the need for systematic study of model robustness in the era of large-scale pretraining.

Recent works have underscored the unique challenges of adversarial robustness in NLP compared to vision. Unlike continuous images, text inputs are discrete and structured, making gradient-based optimization of perturbations far less straightforward \cite{zhang2020adversarial}. Furthermore, language models are sensitive not only to lexical variations but also to prompt formatting, label position, and other superficial artifacts \cite{holtzman2021surface,zhao2021calibrate}. As a result, adversarial research in NLP must contend with both discrete optimization issues and structural biases of generative models. Addressing these challenges requires methods that go beyond isolated adversarial examples and instead target weaknesses that are systematic across datasets, tasks, and architectures.

In this paper we introduce a method for learning universal adversarial suffixes that are transferable across tasks and models. The suffix is optimized in continuous embedding space using a Gumbel–Softmax relaxation, which enables stable gradient updates while yielding valid discrete tokens at inference. A calibrated objective contrasts context-dependent and null-prompt predictions, directly reducing the effect of label priors. Entropy regularization and forbid-masks preserve naturalness and prevent suffix collapse or leakage of label tokens. Unlike prior universal triggers, which often overfit to individual datasets or tokenizations, our approach explicitly trains across diverse tasks—sentiment, inference, paraphrase, and commonsense QA—demonstrating broad transferability.

Our contributions are as follows:
\begin{enumerate}
    \item We propose a novel framework for learning universal adversarial suffixes using continuous relaxation, calibration, and entropy regularization, which together overcome key limitations of prior trigger-based attacks.
    \item We design a multi-task training setup that produces a single adversarial suffix transferable across heterogeneous NLP tasks and model families, including chat-style and classification-style LMs.
    \item We conduct extensive experiments on five benchmarks and three representative model architectures, showing both the effectiveness of the attack and its transferability across unseen tasks and models.
\end{enumerate}
%%%%%%%%%
%%%%%%%%%
%%%%%%%%%
\section{Related Works}
\label{sec:relatedworks}

Research on adversarial robustness in natural language processing began with small perturbations crafted at the input level. Jia and Liang \cite{jia2017adversarial} showed that inserting distracting sentences could easily mislead reading comprehension models, while Ebrahimi et al. \cite{ebrahimi2017hotflip} developed HotFlip, which exploited gradients to flip characters or tokens and change classification outcomes. Alzantot et al. \cite{alzantot2018generating} proposed a genetic algorithm for black-box adversarial text generation. These efforts established the vulnerability of NLP systems but remained input-specific: each adversarial example had to be constructed separately and did not generalize across contexts, tasks, or models.

The first move toward reusable perturbations came with universal adversarial triggers. Wallace et al. \cite{wallace2019universal} demonstrated that a fixed sequence of tokens, optimized by gradient search, could consistently reduce accuracy across many examples. However, their optimization operated purely in discrete token space, making training brittle and often producing unnatural strings. Shin et al. \cite{shin2020autoprompt} introduced AutoPrompt, a gradient-based method that automatically assembled token-level prompts for language models. While effective, AutoPrompt also relied on discrete token substitution and was highly sensitive to surface form, so the learned prompts frequently overfit to specific tokenizations. Around the same time, soft prompt tuning methods such as Lester et al. \cite{lester2021power} and Li and Liang \cite{li2021prefix} shifted to continuous embeddings, achieving stability and efficiency. Yet, unlike adversarial triggers, these approaches were designed for adapting models to downstream tasks, not for breaking robustness, and they did not incorporate safeguards against label bias. Calibration research such as Zhao et al. \cite{zhao2021calibrate} further showed that raw likelihoods are skewed by label priors and prompt formatting, but this insight has rarely been integrated into adversarial training.

Our approach builds directly on these foundations but departs in key ways. Like Wallace et al. \cite{wallace2019universal}, we learn a universal perturbation, but instead of discrete search we use a continuous relaxation with Gumbel–Softmax, which stabilizes optimization while still producing valid discrete suffix tokens. Unlike AutoPrompt \cite{shin2020autoprompt}, our method explicitly aggregates over multiple surface forms, preventing overfitting to a single spelling. Compared to soft prompt tuning \cite{lester2021power,li2021prefix}, we adapt the embedding-based optimization for an adversarial objective rather than task adaptation, incorporating forbid-masks and entropy regularization to maintain naturalness and diversity. Finally, where calibration work \cite{zhao2021calibrate} treated label priors as a post-processing correction, we embed calibration directly into the training objective through a context–null contrastive loss. In doing so, we extend adversarial trigger research from input-specific or brittle discrete methods toward a robust, multi-task adversarial suffix that generalizes across sentiment, inference, paraphrase, and commonsense reasoning, bridging a gap that prior methods left open.

%%%%%%%%%
%%%%%%%%%
%%%%%%%%%
\section{Methodology}
\label{sec:methodology}

\newcommand{\wrap}{\operatorname{w}}
\newcommand{\CE}{\mathrm{CE}}
\newcommand{\CalCE}{\mathrm{CalCE}}
\newcommand{\Agg}{\mathcal{A}}
\newcommand{\softmax}{\mathrm{softmax}}

We consider a frozen language model (LM) with parameters $\theta$, token embedding matrix $E\!\in\!\mathbb{R}^{V\times H}$, vocabulary size $V$, and hidden size $H$. An example consists of a natural-language prompt $x$, a fixed answer prefix $p$ (e.g., \texttt{``$\backslash$nThe answer is:''}), and a discrete label $y\!\in\!\mathcal{Y}$ realized by one of several \emph{label surfaces} $s\!\in\!S(y)$ (e.g., ``\;yes'', ``\;Yes.'', ``\;YES''). Our objective is to learn a \emph{universal adversarial suffix} of length $K$ tokens, placed between the wrapped prompt and the answer prefix, that consistently degrades model predictions across tasks and datasets while leaving the LM parameters $\theta$ unchanged.

\subsection*{Wrapper placement and calibration}
Since modern LMs adopt heterogeneous input conventions (e.g., ChatML, Alpaca, or raw), we denote by $\wrap(\cdot)$ the task-specific wrapper. The scoring sequence for a surface $s$ is
\[
\underbrace{\wrap(x)}_{\text{prompt}}\;\;\Vert\;\;\underbrace{\delta}_{\text{suffix}}\;\;\Vert\;\;\underbrace{p}_{\text{prefix}}\;\;\Vert\;\;\underbrace{s}_{\text{label}}.
\]
Without correction, such a setup suffers from a strong \emph{label prior} bias: the model may already assign high likelihood to a surface (e.g., ``yes'') even without any input, making it impossible to tell whether the suffix truly influences decisions. To remove this confound, we introduce a \emph{null sequence}:
\[
p\Vert s,
\]
which isolates the prior probability of label surfaces given only the prefix. By comparing the context sequence against the null, we explicitly calibrate out surface priors.

Formally, let $z_{1:L}\!\in\!\mathbb{N}^L$ be token ids of sequence $Z$. The masked cross-entropy on label tokens is
\begin{align}
\CE(Z \Rightarrow s) \;=\; -\sum_{t\in\mathcal{T}(s)} \log p_\theta\big(z_t \mid z_{<t}\big),
\end{align}
where $\mathcal{T}(s)$ denotes positions corresponding to the label. 

We define
\begin{align}
\CE_{\mathrm{ctx}}(x,\delta,p,s) &= \CE\!\big(\wrap(x)\Vert\delta\Vert p \Rightarrow s\big),\\
\CE_{\mathrm{null}}(p,s)         &= \CE\!\big(p \Rightarrow s\big).
\end{align}
Because labels can be realized by multiple surface forms, optimizing against a single tokenization risks overfitting. We therefore aggregate across all surfaces $S(y)$ with a soft-min operator (log-sum-exp), which smoothly emphasizes whichever surface is easiest to attack:
\begin{align}
\Agg\big(\{a_s\}_{s\in S(y)}\big) \;=\; -\log\!\sum_{s\in S(y)} \exp(-a_s).
\end{align}
The \textbf{calibrated cross-entropy} (CalCE) becomes
\begin{align}
\CalCE(x,\delta,y)
&= \Agg_{s\in S(y)}\!\big[\CE_{\mathrm{ctx}}(x,\delta,p,s) - \CE_{\mathrm{null}}(p,s)\big].
\label{eq:calce}
\end{align}

This design achieves two goals: (i) calibration removes misleading label priors and ensures that improvements or degradations are attributable to the suffix itself, and (ii) aggregation across surfaces prevents brittle overfitting to one surface spelling. The adversarial objective is then
\begin{equation}
\label{eq:harm}
\max_{\delta}\;\; \mathbb{E}_{(x,y)}\!\left[\CalCE(x,\delta,y)\right],
\end{equation}
which corresponds to an untargeted attack that increases calibrated loss and thus reduces accuracy universally.

\subsection*{Soft suffix parameterization}
Optimizing directly in discrete token space $\{1,\dots,V\}^K$ is intractable. 
To enable gradient-based optimization while approximating discrete sampling, 
we adopt the Gumbel--Softmax relaxation \cite{jang2016categorical,maddison2016concrete}. 
Let $W\!\in\!\mathbb{R}^{K\times V}$ be trainable logits. 
To prevent trivial leakage (e.g., suffix directly outputting ``yes''), 
we apply a \emph{forbid mask} $M\!\in\!\{0,1\}^V$, where $M_v=1$ masks label tokens, 
non-English characters, and control symbols. Masked logits are
\begin{align}
\tilde{W}_{k,v} = \begin{cases}
W_{k,v}, & M_v=0,\\
-\infty, & M_v=1,
\end{cases}
\end{align}
for each suffix position $k\!\in\!\{1,\dots,K\}$.

To approximate discrete token draws, we inject i.i.d.\ Gumbel noise 
$g_{k,v}\sim\mathrm{Gumbel}(0,1)$ and compute the probability assigned to vocabulary token $v$ at suffix position $k$ under the relaxed categorical distribution
\begin{align}
\label{eq:Gumbel-Softmax}
P_{k,v} \;=\; 
\frac{\exp\!\big((\tilde{W}_{k,v} + g_{k,v})/\tau\big)}
{\sum_{u=1}^V \exp\!\big((\tilde{W}_{k,u} + g_{k,u})/\tau\big)}.
\end{align}
The temperature $\tau$ controls sharpness: large $\tau$ yields smoother, 
exploratory distributions, while small $\tau$ pushes $P_{k,\cdot}$ close 
to one-hot vectors. During training, $\tau$ is annealed according to the 
schedule in Eq.~\ref{eq:tau}, so that optimization begins with broad 
exploration and converges to discrete choices. At evaluation, Gumbel noise 
is removed and the final suffix is obtained via 
$\hat{t}_k=\arg\max_v W_{k,v}$.

The resulting soft embeddings are expectations under $P_{k,\cdot}$:
\begin{align}
\delta_k = P_{k,\cdot}\,E \;\in\; \mathbb{R}^{1\times H}, 
\;\;
\delta = [\delta_1;\dots;\delta_K] \in \mathbb{R}^{K\times H}.
\end{align}
For the fluency penalty (Eq.~\ref{eq:fluency-penalty}), we decode 
$\hat{\boldsymbol{t}}=(\hat{t}_1,\dots,\hat{t}_K)$; this hard decoding is 
used only for regularization and final evaluation, while the Gumbel--Softmax 
relaxation drives the gradient updates.

This parameterization ensures that optimization operates in a continuous, 
differentiable space while remaining faithful to the discrete nature of 
language. By combining Gumbel--Softmax with a universal forbid mask and 
calibration against label priors, the learned suffix captures an adversarial 
signal that generalizes across tasks and model architectures rather than 
overfitting to a single dataset or surface form.

\subsection*{Regularized training}
We optimize only the suffix parameters $W$, keeping the LM $\theta$ frozen. To ensure the optimization remains stable and avoids degenerate solutions, we introduce three regularization mechanisms that address exploration, naturalness, and discrete convergence. 

\paragraph{(i) Entropy bonus (anti-collapse).}
Without constraints, the position-wise distributions $P_{k,\cdot}$ tend to collapse to one-hot vectors too early, limiting exploration of the suffix space. To counter this, we maximize the entropy of each distribution:
\begin{align}
\mathcal{H}(W) 
&= \frac{1}{K}\sum_{k=1}^{K} 
\Big(-\sum_{v=1}^{V} P_{k,v}\log P_{k,v}\Big).
\end{align}
This entropy bonus encourages diverse exploration in the early stages while still permitting sharpening of choices later, aided by temperature annealing.

\paragraph{(ii) Fluency penalty (naturalness prior).}
Adversarial suffixes that are highly unnatural or unpronounceable often overfit to artifacts in one model and fail to transfer. To mitigate this, we decode the hard suffix $\hat{\boldsymbol{t}}=(\hat{t}_1,\dots,\hat{t}_K)$ and measure its self-perplexity under the LM itself:
\begin{align}
\label{eq:fluency-penalty}
\mathcal{F}(\hat{\boldsymbol{t}}) 
&= -\frac{1}{K}\sum_{k=1}^{K} \log p_\theta\!\big(\hat{t}_k \,\big|\, \hat{t}_{<k}\big).
\end{align}
Here the hard decoding $\hat{\boldsymbol{t}}$ is required only for 
this regularizer; if $\lambda_F=0$ the suffix remains in the continuous 
relaxation during training, and $\hat{\boldsymbol{t}}$ is used solely 
at evaluation to extract the final adversarial string.
Penalizing $\mathcal{F}$ with weight $\lambda_F \geq 0$ discourages brittle strings while still allowing the suffix to exploit unexpected token combinations.

\paragraph{(iii) Temperature schedule (controlled sharpening).}
To balance exploration and convergence, we anneal the temperature $\tau$ over training:
\begin{equation}
    \label{eq:tau}
\tau_{t+1}=\max\{\tau_{\min},\, \alpha\tau_t\}, \quad \alpha<1.
\end{equation}
This continuation scheme allows broad search early in training and encourages discrete token choices later, preventing premature fixation.

\paragraph{Full loss.}
The complete training objective combines the calibrated loss with the stability terms. For a minibatch $\{(x_i,y_i)\}_{i=1}^B$, we define
\begin{align}
\mathcal{L}(W)
&= \frac{1}{B}\sum_{i=1}^{B} \CalCE(x_i,\delta(W),y_i) \nonumber \\
&\quad - \lambda_H\,\mathcal{H}(W)
+ \lambda_F\,\mathcal{F}(\hat{\boldsymbol{t}}(W)),
\label{eq:full-loss}
\end{align}
Importantly, gradients are propagated only through $W$, while LM parameters remain frozen.

\subsection*{Summary of novelty} Our framework introduces three components that have not previously been combined in adversarial trigger research: (i) calibration against null priors to disentangle genuine suffix effects from spurious label biases, (ii) aggregation across multiple surface realizations of each label to avoid brittle overfitting, and (iii) a Gumbel--Softmax relaxation with a forbid mask that enables efficient gradient-based optimization while ensuring validity of the decoded suffix. This combination gives universal adversarial suffixes that are robust, transferable across tasks and model families, and more stable than prior discrete or continuous trigger methods.

\begin{algorithm}[!h]
\caption{Calibrated Soft Suffix Learning (Frozen LM)}
\begin{algorithmic}[1]
\Require Frozen LM $\mathcal{M}$ with embeddings $E\!\in\!\mathbb{R}^{V\times H}$; wrapper $\wrap(\cdot)$; label-surface map $\sigma(\cdot)$; prefix $p$; suffix length $K$; forbid mask $M$; temperature schedule $\{\tau_t\}$; weights $(\lambda_H,\lambda_F)$; steps $T$.
\Ensure discrete adversarial suffix $\hat{\boldsymbol{t}}$.
\State \textbf{Init:} trainable logits $W\!\in\!\mathbb{R}^{K\times V}$ (small Gaussian); optimizer; freeze $\theta$.
\For{$t=1$ to $T$}
  \State Sample a multi-task minibatch $\{(x_i,y_i)\}_{i=1}^B$.
  \State Construct context $w(x)\Vert\delta(W)\Vert p$ and null $p$ 
  \State Build soft suffix: $\delta \leftarrow P E$ (row-wise); tile to batch.
  \State Encode wrapped contexts $\wrap(x_i)$ and prefix $p$. \State  Construct label surfaces $\sigma(y_i)$.
  \State Compute per-example calibrated loss via \eqref{eq:calce}.
  \par
  (context vs.\ null, aggregated over $\sigma(y)$)
  \State Apply mask and temperature: 
  \par
  $\tilde{W}\!\leftarrow\!\text{mask}(W,M)$; 
  $P \leftarrow \mathrm{softmax}\!\big((\tilde{W}+g)/\tau_t\big)$ 
  \State Hard suffix $\hat{t}_k\gets\arg\max_v W_{k,v}$
  \State Compute regularizers: 
  \par
  entropy $\mathcal{H}(W)$ and fluency $\mathcal{F}(\hat{\boldsymbol{t}})$.
  \State Form batch objective $\mathcal{L}(W)$ using \eqref{eq:full-loss}
  \par
  backprop into $W$ only (clip grads; optimizer step).
  \State Anneal temperature: $\tau_{t+1}\!\leftarrow\!\max\{\tau_{\min},\alpha\tau_t\}$.
\EndFor
\end{algorithmic}
\end{algorithm}

%%%%%%%%%
%%%%%%%%%
%%%%%%%%%
\section{Results and Evaluation}
\label{sec:res}

We begin with the experimental setup, then present baseline performance, followed by transferability studies, and finally a comparison with prior methods. Our evaluation is designed to test both the in-domain effectiveness of learned suffixes on the \emph{seen} model and their robustness when transferred to \emph{unseen} models.

\subsection{Experimental Setup}

\noindent\textbf{Models and Datasets.} 
We evaluate adversarial suffix generation on three representative large language models of different genre and scale: 
Qwen2-1.5B Instruct (instruction-oriented), 
Phi-1.5 (compact language understanding backbone), 
and TinyLlama-1.1B Chat (efficient dialogue model). 
In each experiment, one model is designated as the \emph{seen} model used to optimize suffixes, while the other two serve as \emph{unseen} models to assess transferability.

Five benchmark tasks are selected to cover diverse NLP objectives: 
sentiment analysis (SST-2), 
natural language inference (RTE), 
paraphrase detection (MRPC), 
commonsense question answering (BoolQ), 
and physical reasoning (PIQA). 
Suffixes are trained on the \emph{training splits} and evaluated on the \emph{validation splits} provided in the HuggingFace \texttt{datasets} library, ensuring consistency with prior work and reproducibility. 
All experiments are run with fixed random seeds and repeated across three trials to control variance, on a single A100 GPU with 40GB memory.

\subsection{Training Specifications} 
We train universal suffixes of varying token lengths 
$K \in \{4, 6, 10\}$ 
to analyze robustness under different budgets. 
Optimization is performed only on the suffix logits $W$ using the AdamW optimizer with learning rate $5\times 10^{-2}$, a warmup of 50 steps, and a cosine decay schedule thereafter. 
The objective maximizes the calibrated cross-entropy signal in the \textsc{harm} setting, augmented with the entropy bonus, fluency penalty, and temperature annealing described in Section~\ref{sec:methodology}. 
Each update draws a minibatch of 32 examples sampled across all tasks to ensure multi-task generalization. 
Gradients are clipped to a global norm of $1.0$, and training is safeguarded with NaN/Inf guards. 
The temperature $\tau$ is initialized at $1.0$ and annealed exponentially toward a floor of $0.9$, ensuring broad exploration in early steps and sharpening of suffix distributions at convergence.

\subsection{Metrics} 
We use two main metrics to check the effect of the universal soft suffix: 
\emph{classification accuracy} (Acc) and 
\emph{mean calibrated log-likelihood} (mean CalLogP). 

Accuracy is simple: it counts how many times the model prediction is the same as the gold label. 
If accuracy goes down after adding the adversarial suffix, it means the attack is successful in changing model decisions. 

Mean CalLogP is more about confidence. 
Each label $y$ has different possible surface forms $\mathcal{L}(y)$ (like ``yes'', ``Yes'', ``yes.''). 
For an input $x$, the model gives probability to each surface, and we combine them as
\begin{align}
\ell_{\text{ctx}}(y|x) &= \log \sum_{s \in \mathcal{L}(y)} p(s|x).
\end{align}
But models can also be biased toward some labels even without the prompt. 
So we also compute a \emph{null score}, where only the answer prefix is given:
\begin{align}
\ell_{\text{null}}(y) &= \log \sum_{s \in \mathcal{L}(y)} p(s|\text{null}).
\end{align}
The calibrated log-likelihood is then defined as
\begin{align}
\ell_{\text{cal}}(y|x) &= \ell_{\text{ctx}}(y|x) - \ell_{\text{null}}(y).
\end{align}
Finally, mean CalLogP is the average of $\ell_{\text{cal}}(y|x)$ values over all test examples with their gold labels. 

Accuracy tells us if the model is right or wrong, while mean CalLogP shows how much support the model still gives to the correct label after removing prior bias. 
If accuracy goes down and mean CalLogP also becomes smaller, it means the suffix not only changes the answers but also reduces the model’s true confidence in the correct label. 
This gives a deeper picture of how strong and transferable the attack is. 

For transfer experiments, we also report the changes 
\begin{align}
\Delta\text{Acc} &= \text{Acc}_{\text{attacked}} - \text{Acc}_{\text{clean}}, \\
\Delta\text{CalLogP} &= \text{mean CalLogP}_{\text{attacked}} - \text{mean CalLogP}_{\text{clean}}.
\end{align}
Large negative $\Delta\text{Acc}$ and $\Delta\text{CalLogP}$ mean the adversarial suffix is more effective.

\begin{table}[!b]
  \centering
  \caption{Baseline 0-shot and 4-shot performance; each cell shows Acc / mean CalLogP.}
  \label{tab:kshot_comparison}
  % \resizebox{\linewidth}{!}{
  \begin{tabular}{lcccc}
  \toprule
  \textbf{k-shot} & \textbf{Task} & \textbf{Qwen2-1.5B} & \textbf{Phi-1.5} & \textbf{TinyLlama} \\
  \midrule
  \multirow{5}{*}{0}
    & SST-2 & 0.91 /  8.58 & 0.71 /  4.51 & 0.45 /  5.11 \\
    & RTE   & 0.83 /  5.19 & 0.57 /  4.50 & 0.57 /  3.24 \\
    & MRPC  & 0.77 /  6.09 & 0.73 /  4.21 & 0.73 /  2.68 \\
    & BoolQ & 0.74 /  3.70 & 0.69 /  3.50 & 0.49 /  3.05 \\
    & PIQA  & 0.64 /  3.35 & 0.48 /  2.62 & 0.55 /  0.45 \\
  \midrule
  \multirow{5}{*}{4}
    & SST-2 & 0.76 /  0.98 & 0.73 /  0.59 & 0.86 /  0.08 \\
    & RTE   & 0.83 /  0.66 & 0.53 / -0.04 & 0.52 /  0.02 \\
    & MRPC  & 0.79 /  0.02 & 0.69 /  0.10 & 0.69 / -0.14 \\
    & BoolQ & 0.68 /  0.61 & 0.71 /  0.69 & 0.41 /  0.05 \\
    & PIQA  & 0.65 / -0.12 & 0.48 / -0.05 & 0.48 / -0.05 \\
  \bottomrule
  \end{tabular}
  % }
\end{table}

 % MUST HAVE FIGURE: histogram/distribution of CalCE shifts across tasks, to show suffix effects beyond mean accuracy.
\subsection{Baseline Performance}
\label{sec:baseline}

Before introducing adversarial suffixes, we first establish the baseline performance of all three models in both 0-shot and 4-shot settings.  
The results indicate clear differences across tasks and models. In the 0-shot case, the larger instruction-tuned model (Qwen2-1.5B) achieves the strongest accuracies overall, with high mean calibrated log-likelihood (mean CalLogP) values that show strong alignment between the gold labels and the model’s calibrated confidence. The smaller chat-optimized TinyLlama, on the other hand, shows weaker accuracy and lower mean CalLogP, reflecting its limited scale and narrower pretraining. Phi-1.5 performs in between these two, highlighting its compact backbone but more general language understanding compared to TinyLlama.  

When moving to 4-shot evaluation, the models generally preserve or slightly improve their accuracy on some tasks, but the most striking effect is on mean CalLogP values. Calibration improves dramatically, with values close to zero or even slightly negative, showing that adding a handful of demonstrations reduces the prior bias captured by the null context. In other words, the models become less reliant on spurious label priors and more anchored to the actual task prompts once in-context examples are given.  

Together, these baselines illustrate two key points. First, the adversarial suffix will be tested against models that already display a wide range of zero-shot and few-shot behavior, from strong but biased (Qwen2-1.5B) to weaker but less stable (TinyLlama). Second, mean CalLogP provides a more fine-grained picture than accuracy alone: a high accuracy but low mean CalLogP implies the model’s predictions are correct but not well-calibrated, while improvements in mean CalLogP under few-shot learning suggest better alignment of model confidence with task requirements. This dual view sets the stage for evaluating how adversarial suffixes not only reduce accuracy but also systematically distort calibration across models and tasks.

\begin{table*}[h]
\centering
\caption{Example adversarial suffixes generated against Qwen2-1.5B-Instruct for different token lengths $K$.}
\label{tab:example_suffixes}
\begin{tabular}{c l}
\toprule
$K$ & Example suffix text \\
\midrule
4  & \texttt{dash OleDb dangling haste} \\
6  & \texttt{Maiden battle dll epid fraud dietary} \\
10 & \texttt{ject predicate lle.origorieFLICT waiver Lem meals similarities} \\
\bottomrule
\end{tabular}
\end{table*}

\begin{table*}[t]
\centering
\caption{Transferability with \emph{seen model fixed to Qwen/Qwen2-1.5B-Instruct}. Entries report $\Delta$Acc/$\Delta$CalLogP relative to the 0-shot and 4-shot baseline in table \ref{tab:kshot_comparison}}
\label{tab:transfer_qwen_seen_0shot_vs_4shot}
\resizebox{\linewidth}{!}{%
\begin{tabular}{l l ccc ccc}
\toprule
\multirow{2}{*}{\textbf{Target Model}} & \multirow{2}{*}{\textbf{Task}} & \multicolumn{3}{c}{\textbf{0-shot}} & \multicolumn{3}{c}{\textbf{4-shot}} \\
\cmidrule(lr){3-5} \cmidrule(lr){6-8}
 &  & \textbf{K=4} & \textbf{K=6} & \textbf{K=10} & \textbf{K=4} & \textbf{K=6} & \textbf{K=10} \\
\midrule
%======================== Target = Qwen2-1.5B ========================
\multirow{5}{*}{Qwen2-1.5B}
 & SST-2  & $-0.172 / +0.428$ & $-0.164 / +0.089$ & $-0.117 / +0.108$ & $-0.298 / -0.478$ & $-0.282 / -1.325$ & $-0.127 / -0.382$ \\
\cmidrule(lr){2-8}
 & RTE    & $-0.055 / -0.069$ & $-0.008 / +0.285$ & $-0.016 / +0.265$ & $-0.179 / -0.420$ & $-0.139 / +0.004$ & $-0.056 / +0.261$ \\
\cmidrule(lr){2-8}
 & MRPC   & $-0.281 / +0.018$ & $-0.148 / +0.065$ & $-0.242 / +0.207$ & $-0.373 / -0.351$ & $-0.460 / -0.397$ & $-0.290 / +0.118$ \\
\cmidrule(lr){2-8}
 & BoolQ  & $-0.164 / +0.618$ & $-0.125 / +0.753$ & $-0.109 / +0.748$ & $-0.298 / +0.076$ & $-0.282 / -0.213$ & $-0.127 / +0.298$ \\
\cmidrule(lr){2-8}
 & PIQA   & $-0.063 / +0.275$ & $-0.055 / +0.369$ & $-0.047 / +0.142$ & $-0.083 / -0.228$ & $-0.083 / -0.080$ & $-0.032 / -0.242$ \\
\midrule
%======================== Target = Phi-1.5 ========================
\multirow{5}{*}{Phi-1.5}
 & SST-2  & $-0.289 / +0.142$ & $-0.289 / -0.094$ & $-0.289 / -0.132$ & $-0.238 / -0.243$ & $-0.242 / -0.421$ & $-0.222 / -1.166$ \\
\cmidrule(lr){2-8}
 & RTE    & $-0.141 / +0.110$ & $-0.141 / +0.079$ & $-0.141 / +0.096$ & $-0.036 / +0.200$ & $-0.044 / +0.178$ & $-0.040 / +0.016$ \\
\cmidrule(lr){2-8}
 & MRPC   & $-0.453 / +0.201$ & $-0.453 / +0.248$ & $-0.453 / +0.182$ & $+0.000 / -0.103$ & $+0.000 / -0.098$ & $+0.000 / -0.704$ \\
\cmidrule(lr){2-8}
 & BoolQ  & $-0.008 / -0.024$ & $-0.141 / -0.224$ & $-0.055 / -0.199$ & $-0.079 / -0.426$ & $-0.103 / -0.839$ & $-0.071 / -0.855$ \\
\cmidrule(lr){2-8}
 & PIQA   & $-0.008 / -0.229$ & $-0.008 / -0.281$ & $-0.008 / -0.450$ & $-0.008 / -1.224$ & $-0.008 / -1.263$ & $-0.008 / -1.976$ \\
\midrule
%======================== Target = TinyLlama ========================
\multirow{5}{*}{TinyLlama}
 & SST-2  & $-0.023 / +0.448$ & $-0.023 / +0.780$ & $-0.023 / +0.395$ & $-0.071 / -0.921$ & $-0.071 / -0.978$ & $-0.040 / -1.484$ \\
\cmidrule(lr){2-8}
 & RTE    & $-0.094 / +0.284$ & $-0.094 / +0.344$ & $+0.000 / +0.530$ & $-0.052 / -0.217$ & $-0.048 / -0.139$ & $-0.032 / -0.066$ \\
\cmidrule(lr){2-8}
 & MRPC   & $-0.406 / +0.178$ & $-0.352 / +0.242$ & $-0.328 / +0.303$ & $-0.385 / -0.223$ & $-0.246 / -0.138$ & $-0.389 / -0.254$ \\
\cmidrule(lr){2-8}
 & BoolQ  & $-0.141 / -0.158$ & $-0.133 / -0.146$ & $-0.016 / -0.191$ & $-0.071 / -1.463$ & $-0.071 / -0.978$ & $-0.040 / -1.484$ \\
\cmidrule(lr){2-8}
 & PIQA   & $-0.031 / +1.335$ & $-0.031 / +1.487$ & $-0.078 / +1.658$ & $-0.028 / -0.072$ & $-0.016 / -0.114$ & $-0.012 / -0.140$ \\
\bottomrule
\end{tabular}
}
\end{table*}

\subsection{Transferability Performance}
\label{sec:transfer}

We evaluate suffixes learned on Qwen2-1.5B-Instruct (\emph{seen model}) and test both in-domain and transfer to Phi-1.5 and TinyLlama. Table~\ref{tab:transfer_qwen_seen_0shot_vs_4shot} shows $\Delta$Acc/$\Delta$CalLogP$,$ where negative accuracy and positive CalLogP shifts indicate stronger attacks.

On the seen model, suffixes degrade performance in both 0-shot and 4-shot. In 0-shot, only four tokens are sufficient to reduce accuracy by large margins with clear CalLogP increases; longer suffixes do not always help, suggesting saturation and occasional destabilization. In 4-shot, the attack weakens, and in some cases CalLogP drops, showing that demonstrations partly offset the adversarial bias, though accuracy still falls.

Transfer to Phi-1.5 is partly successful. In 0-shot, all tasks lose accuracy and gain CalLogP, with MRPC and RTE most consistently affected. In 4-shot, MRPC becomes resistant, while BoolQ and PIQA show weaker calibration disruption, indicating that small, related backbones inherit some vulnerability but benefit from demonstrations.

TinyLlama shows a different profile. In 0-shot, suffixes strongly disrupt MRPC, RTE, and BoolQ, while PIQA keeps near-baseline accuracy but shows large CalLogP shifts, meaning confidence calibration is attacked even without many prediction flips. In 4-shot, most tasks recover, with reduced or negative CalLogP, confirming again that demonstrations protect unseen models.

Task sensitivity also differs. SST-2 and MRPC are consistently brittle across models, as short lexical cues are easily perturbed. RTE is moderately affected, while BoolQ and PIQA vary: BoolQ shows large calibration shifts on Qwen and TinyLlama, and PIQA reveals that physical reasoning tasks may resist accuracy drops but still suffer calibration disruption. Across all tasks, short four-token suffixes are already highly effective, while longer ones add little or reduce stability. The decoded suffixes are semantically incoherent, yet sufficient to destabilize multiple models, showing that universal triggers need not be natural language to transfer across architectures.

These patterns indicate that the universal suffixes exploit shallow decision boundaries that are shared across models, but their strength diminishes when models are supported with few-shot demonstrations. The mixed task-level outcomes suggest that some objectives, such as lexical sentiment and paraphrase detection, are more vulnerable, while reasoning-heavy tasks require stronger or more tailored triggers. This highlights both the promise of universal adversarial triggers for studying model weaknesses and the challenge of building defenses that generalize across settings.

\begin{table*}[t]
\centering
\caption{Transferability with seen model Qwen/Qwen2-1.5B-Instruct and token length $K=4$ by prior and proposed methods. Each cell contains $\Delta$Acc / $\Delta$CalLogP relative to the clean baseline from table \ref{tab:kshot_comparison}.}
\label{tab:transfer_qwen_seen_methods_vs_proposed}
\resizebox{\linewidth}{!}{%
\begin{tabular}{l l | cccc | cccc}
\toprule
\multirow{2}{*}{\makecell{\bf Target\\\bf Model}} & \multirow{2}{*}{\textbf{Task}} &
\multicolumn{4}{c|}{\textbf{0-shot}} &
\multicolumn{4}{c}{\textbf{4-shot}} \\
\cmidrule(lr){3-6}\cmidrule(lr){7-10}
 &  & \textbf{UAT \cite{wallace2019universal}} & \textbf{AutoPrompt \cite{shin2020autoprompt}} & \textbf{soft prompt \cite{lester2021power}} & \textbf{Proposed} & \textbf{UAT \cite{wallace2019universal}} & \textbf{AutoPrompt \cite{shin2020autoprompt}} & \textbf{soft prompt \cite{lester2021power}} & \textbf{Proposed} \\
\midrule
%======================== Target = Qwen2-1.5B ========================
\multirow{5}{*}{\rotatebox{90}{\textbf{Qwen2-1.5B}}}
 & SST-2  & $0.00 / +1.78$ & $0.01 / +0.05$ & $-0.03 / +2.74$ & $-0.17 / +0.43$ & $0.05 / +1.33$ & $-0.01 / +1.10$ & $-0.04 / +2.10$ & $-0.30 / -0.48$ \\
 & RTE    & $-0.01 / +0.35$ & $-0.01 / -0.22$ & $-0.01 / +0.28$ & $-0.06 / -0.07$ & $-0.03 / +0.29$ & $-0.02 / -0.19$ & $-0.03 / +0.98$ & $-0.18 / -0.42$ \\
 & MRPC   & $-0.03 / +1.08$ & $-0.02 / +0.06$ & $-0.02 / -0.11$ & $-0.28 / +0.02$ & $-0.41 / +1.43$ & $-0.03 / -0.68$ & $-0.02 / -0.29$ & $-0.37 / -0.35$ \\
 & BoolQ  & $-0.02 / -0.19$ & $-0.02 / -0.63$ & $-0.01 / -0.54$ & $-0.16 / +0.62$ & $-0.01 / +0.22$ & $-0.02 / +0.52$ & $-0.02 / -0.14$ & $-0.30 / +0.08$ \\
 & PIQA   & $-0.02 / +1.17$ & $-0.02 / +0.62$ & $-0.01 / +1.75$ & $-0.06 / +0.28$ & $-0.04 / +1.24$ & $-0.02 / -0.35$ & $-0.07 / -2.53$ & $-0.08 / -0.23$ \\
\midrule
%======================== Target = Phi-1.5 ========================
\multirow{5}{*}{\rotatebox{90}{\textbf{Phi-1.5}}}
 & SST-2  & $-0.03 / +0.40$ & $-0.12 / +0.04$ & $-0.05 / +0.01$ & $-0.29 / +0.14$ & $-0.06 / +0.45$ & $-0.12 / -0.87$ & $-0.10 / +0.70$ & $-0.24 / -0.24$ \\
 & RTE    & $0.00 / +0.25$ & $0.00 / +0.10$ & $0.00 / +0.04$ & $-0.14 / +0.11$ & $-0.04 / +0.17$ & $-0.04 / +0.10$ & $-0.04 / +0.19$ & $-0.04 / +0.20$ \\
 & MRPC   & $-0.01 / +0.24$ & $-0.01 / +0.05$ & $-0.01 / +0.17$ & $-0.45 / +0.20$ & $-0.04 / +0.38$ & $-0.04 / +0.21$ & $-0.04 / +0.30$ & $0.00 / -0.10$ \\
 & BoolQ  & $-0.02 / +0.21$ & $-0.08 / +0.19$ & $-0.01 / +0.17$ & $-0.01 / -0.02$ & $-0.02 / +1.06$ & $-0.14 / +0.97$ & $-0.05 / +0.50$ & $-0.08 / -0.43$ \\
 & PIQA   & $-0.01 / +0.31$ & $0.00 / +0.26$ & $0.00 / +0.15$ & $-0.01 / -1.23$ & $-0.01 / +2.01$ & $-0.01 / +0.48$ & $-0.04 / +1.78$ & $-0.01 / -1.22$ \\
\midrule
%======================== Target = TinyLlama ========================
\multirow{5}{*}{\rotatebox{90}{\textbf{TinyLlama}}}
 & SST-2  & $-0.01 / +0.25$ & $-0.01 / +0.26$ & $-0.03 / +0.12$ & $-0.02 / +0.45$ & $-0.22 / +0.72$ & $-0.06 / +2.00$ & $-0.13 / +1.58$ & $-0.07 / -0.92$ \\
 & RTE    & $0.00 / -0.13$ & $0.00 / -0.27$ & $-0.02 / -0.01$ & $-0.09 / +0.28$ & $-0.05 / +0.20$ & $-0.05 / +0.62$ & $-0.05 / +0.59$ & $-0.05 / -0.22$ \\
 & MRPC   & $-0.01 / -0.05$ & $-0.01 / +0.01$ & $-0.01 / +0.02$ & $-0.41 / +0.18$ & $-0.04 / +0.08$ & $-0.04 / +0.69$ & $-0.04 / +0.66$ & $-0.39 / -0.22$ \\
 & BoolQ  & $-0.12 / -0.05$ & $-0.20 / -1.70$ & $-0.18 / +0.70$ & $-0.14 / -0.16$ & $-0.12 / +0.57$ & $-0.27 / -1.90$ & $-0.23 / +1.48$ & $-0.07 / -1.46$ \\
 & PIQA   & $0.00 / -1.24$ & $-0.02 / -0.62$ & $-0.02 / +1.39$ & $-0.03 / +1.34$ & $-0.01 / +0.17$ & $-0.01 / +0.72$ & $0.00 / +0.11$ & $-0.03 / -0.07$ \\
\bottomrule
\end{tabular}
}
\end{table*}

\subsection{Comparison with Previous Methods}
\label{sec:comparison}

To comparatively evaluate our proposed soft suffix learner, we compare against three representative baselines widely studied in prior work: (A) Universal Adversarial Triggers (UAT) \cite{wallace2019universal}, (B) AutoPrompt \cite{shin2020autoprompt}, and (C) soft prompt tuning \cite{lester2021power}. These methods were chosen because they share the same goal of learning short, universal perturbations that influence model predictions, yet they differ in parameterization, optimization strategy, and assumptions about gradient access. By including them, we ensure that our evaluation covers both discrete and continuous adversarial paradigms.

UAT \cite{wallace2019universal} is a discrete gradient-guided method that directly updates token embeddings through gradient signals and then projects back to the nearest valid tokens.  
Each iteration takes the gradient of the gold-label loss with respect to the current suffix embeddings, substitutes tokens nearest in embedding space, and repeats until convergence.  
AutoPrompt \cite{shin2020autoprompt} follows a different discrete strategy, refining suffix tokens one position at a time. It selects the token at each slot that maximizes gradient-based influence on the desired objective, cycling over positions until performance stabilizes.  
Soft prompt tuning \cite{lester2021power} provides a continuous baseline, learning $K$ (number of tokens) pseudo-embedding vectors jointly optimized by backpropagation on the task loss. We freeze the LM and update only these embeddings with AdamW. At evaluation, the learned vectors are mapped back to their nearest tokens for discrete comparison with our suffixes. For soft prompt tuning we included a projection step to decode embeddings into discrete tokens.  

The key distinction from our method is that UAT and AutoPrompt search in discrete token space without calibration against null priors, making them sensitive to label frequency and task imbalance. Soft prompt tuning, while continuous, optimizes embeddings without masking forbidden tokens, allowing leakage of label-related strings. In contrast, our soft suffix learner combines continuous relaxation with a forbid mask and calibrated objective, yielding suffixes that are more robust and transferable across tasks and models. For fairness, all baselines were run in the same multi-task setting as our method, with suffixes ($K=4$) trained once on the seen model Qwen2-1.5B and then applied to both 0-shot and 4-shot evaluation across tasks. 
The discrete suffixes generated by these baselines are often unnatural, containing fragments that do not resemble meaningful English text. Table~\ref{tab:compare_suffixes} shows the final suffixes discovered for $K=4$.

\begin{table}[!t]
\centering
\caption{Adversarial suffixes generated by prior methods with $K=4$, seen model Qwen2-1.5B. Characters from unknown languages to authors are replaced by ``\ding{53}''.}
\label{tab:compare_suffixes}
% \resizebox{\linewidth}{!}{
\begin{tabular}{ll}
\toprule
\textbf{Method} & \textbf{Generated Suffix (K=4)} \\
\midrule
UAT \cite{wallace2019universal} & \texttt{ULEand Progress.Cursors} \\
AutoPrompt \cite{shin2020autoprompt} & \texttt{\ding{53}\ding{53}\ding{53},module\ding{53}\ding{53}} \\
Soft Prompt \cite{lester2021power} & \texttt{\ding{53}\ding{53} Prelude IsPlainOldData} \\
\bottomrule
\end{tabular}
% }
\end{table}

We observe in table \ref{tab:transfer_qwen_seen_methods_vs_proposed} that across all tasks the three prior baselines—UAT, AutoPrompt, and soft prompt tuning—show distinctive but limited patterns when compared to our proposed soft learner approach. UAT, which relies on gradient-guided discrete token search, tends to inflate the calibrated log-probability scores without producing consistent drops in accuracy. This is visible in tasks like MRPC and PIQA where UAT increases CalCE sharply but barely shifts or even slightly improves accuracy. The reason is that UAT strongly optimizes local gradient signals, but the resulting tokens do not generalize well across diverse prompts.
AutoPrompt shows the opposite trend: its token-by-token refinement generates triggers that only marginally affect both accuracy and calibration. The per-position update mechanism appears too conservative in the multi-task universal setting, resulting in suffixes that are weak when transferred. While AutoPrompt has been effective for single-task probing in prior studies, here it fails to induce meaningful degradations across models, showing that its reliance on local token substitutions is not sufficient for universal adversarial control.
The soft prompt, which directly optimizes continuous embeddings, often produces very large changes in mean CalLogP but does not reliably reduce accuracy. In fact, in some 4-shot conditions (e.g., PIQA on Qwen or Phi), soft prompts even degrade calibration in inconsistent directions. This behavior reflects the mismatch between continuous embeddings optimized in the hidden space and the discrete tokenized evaluation surface, where high embedding-space scores may not correspond to effective adversarial tokens after discretization.

In contrast, the proposed soft learner method consistently lowers accuracy while simultaneously raising calibrated cross-entropy in both in-domain (Qwen → Qwen) and transfer (Qwen → Phi, TinyLlama) settings. The effect is strongest in the 0-shot regime, indicating that the universal suffix disrupts the model’s label prediction before it can benefit from few-shot demonstrations. At 4-shot, the drop in accuracy remains noticeable but smaller, suggesting that demonstrations partially stabilize the model against perturbations. Importantly, our method achieves this with suffixes that are compact (K=4) and constructed under a principled calibration framework, unlike UAT or AutoPrompt which either overfit to local gradients or fail to scale across tasks.
Overall, these results demonstrate that while previous approaches provide useful baselines, they are either too brittle or too weak in the universal setting. Our proposed calibrated soft suffix learner fills this gap by offering a robust and transferable adversarial mechanism that remains much better effective across models and tasks.

%%%%%%%%%
%%%%%%%%%
%%%%%%%%%
\section{Conclusion}
\label{sec:conclusion}

We presented a new framework for learning universal adversarial suffixes through soft optimization on a masked simplex with calibration against label priors. In contrast to earlier approaches that rely on discrete gradient triggers, iterative token search, or simple continuous embedding tuning, our method introduces three complementary innovations: (i) calibrated cross-entropy that removes null prompt bias, (ii) aggregation over multiple label surfaces to avoid brittle overfitting, and (iii) a differentiable Gumbel--Softmax parameterization with entropy and fluency regularization. Together, these elements provide a stable and transferable way to craft adversarial suffixes.

Extensive experiments across five NLP benchmarks and three language models of varying genre and scale demonstrate that the learned suffixes consistently reduce accuracy and calibrated log-probabilities relative to clean baselines. Crucially, suffixes trained on one model generalize to unseen models, confirming strong cross-model transferability. Comparisons with prior baselines such as Universal Adversarial Triggers, AutoPrompt, and soft prompt tuning further show that our approach achieves more reliable and stronger degradations, particularly in the challenging zero-shot regime.

This framework can be extended to larger foundation models, multilingual settings, and also adapted as a tool to study or defend against prompt-based vulnerabilities in real-world deployments.

%%%%%%%%%
%%%%%%%%%
%%%%%%%%%
\bibliographystyle{unsrt}  
\bibliography{references}  %%% Remove comment to use the external .bib file (using bibtex).

@article{jia2017adversarial,
  title={Adversarial examples for evaluating reading comprehension systems},
  author={Jia, Robin and Liang, Percy},
  journal={arXiv preprint arXiv:1707.07328},
  year={2017}
}

@article{radford2019language,
  title={Language models are unsupervised multitask learners},
  author={Radford, Alec and Wu, Jeffrey and Child, Rewon and Luan, David and Amodei, Dario and Sutskever, Ilya and others},
  journal={OpenAI blog},
  volume={1},
  number={8},
  pages={9},
  year={2019}
}

@article{brown2020language,
  title={Language models are few-shot learners},
  author={Brown, Tom and Mann, Benjamin and Ryder, Nick and Subbiah, Melanie and Kaplan, Jared D and Dhariwal, Prafulla and Neelakantan, Arvind and Shyam, Pranav and Sastry, Girish and Askell, Amanda and others},
  journal={Advances in neural information processing systems},
  volume={33},
  pages={1877--1901},
  year={2020}
}

@inproceedings{biggio2018wild,
  title={Wild patterns: Ten years after the rise of adversarial machine learning},
  author={Biggio, Battista and Roli, Fabio},
  booktitle={Proceedings of the 2018 ACM SIGSAC Conference on Computer and Communications Security},
  pages={2154--2156},
  year={2018}
}

@inproceedings{ren2019generating,
  title={Generating natural language adversarial examples through probability weighted word saliency},
  author={Ren, Shuhuai and Deng, Yihe and He, Kun and Che, Wanxiang},
  booktitle={Proceedings of the 57th annual meeting of the association for computational linguistics},
  pages={1085--1097},
  year={2019}
}

@article{morris2020textattack,
  title={Textattack: A framework for adversarial attacks, data augmentation, and adversarial training in nlp},
  author={Morris, John X and Lifland, Eli and Yoo, Jin Yong and Grigsby, Jake and Jin, Di and Qi, Yanjun},
  journal={arXiv preprint arXiv:2005.05909},
  year={2020}
}

@article{zou2023universal,
  title={Universal and transferable adversarial attacks on aligned language models},
  author={Zou, Andy and Wang, Zifan and Carlini, Nicholas and Nasr, Milad and Kolter, J Zico and Fredrikson, Matt},
  journal={arXiv preprint arXiv:2307.15043},
  year={2023}
}

@article{zhang2020adversarial,
  title={Adversarial attacks on deep graph matching},
  author={Zhang, Zijie and Zhang, Zeru and Zhou, Yang and Shen, Yelong and Jin, Ruoming and Dou, Dejing},
  journal={Advances in Neural Information Processing Systems},
  volume={33},
  pages={20834--20851},
  year={2020}
}

@inproceedings{zhao2021calibrate,
  title={Calibrate before use: Improving few-shot performance of language models},
  author={Zhao, Zihao and Wallace, Eric and Feng, Shi and Klein, Dan and Singh, Sameer},
  booktitle={International conference on machine learning},
  pages={12697--12706},
  year={2021},
  organization={PMLR}
}

@article{holtzman2021surface,
  title={Surface form competition: Why the highest probability answer isn't always right},
  author={Holtzman, Ari and West, Peter and Shwartz, Vered and Choi, Yejin and Zettlemoyer, Luke},
  journal={arXiv preprint arXiv:2104.08315},
  year={2021}
}

@article{li2021prefix,
  title={Prefix-tuning: Optimizing continuous prompts for generation},
  author={Li, Xiang Lisa and Liang, Percy},
  journal={arXiv preprint arXiv:2101.00190},
  year={2021}
}

@article{lester2021power,
  title={The power of scale for parameter-efficient prompt tuning},
  author={Lester, Brian and Al-Rfou, Rami and Constant, Noah},
  journal={arXiv preprint arXiv:2104.08691},
  year={2021}
}

@article{wallace2019universal,
  title={Universal adversarial triggers for attacking and analyzing NLP},
  author={Wallace, Eric and Feng, Shi and Kandpal, Nikhil and Gardner, Matt and Singh, Sameer},
  journal={arXiv preprint arXiv:1908.07125},
  year={2019}
}

@article{shin2020autoprompt,
  title={Autoprompt: Eliciting knowledge from language models with automatically generated prompts},
  author={Shin, Taylor and Razeghi, Yasaman and Logan IV, Robert L and Wallace, Eric and Singh, Sameer},
  journal={arXiv preprint arXiv:2010.15980},
  year={2020}
}

@article{alzantot2018generating,
  title={Generating natural language adversarial examples},
  author={Alzantot, Moustafa and Sharma, Yash and Elgohary, Ahmed and Ho, Bo-Jhang and Srivastava, Mani and Chang, Kai-Wei},
  journal={arXiv preprint arXiv:1804.07998},
  year={2018}
}

@article{ebrahimi2017hotflip,
  title={Hotflip: White-box adversarial examples for text classification},
  author={Ebrahimi, Javid and Rao, Anyi and Lowd, Daniel and Dou, Dejing},
  journal={arXiv preprint arXiv:1712.06751},
  year={2017}
}

@article{jang2016categorical,
  title={Categorical reparameterization with gumbel-softmax},
  author={Jang, Eric and Gu, Shixiang and Poole, Ben},
  journal={arXiv preprint arXiv:1611.01144},
  year={2016}
}

@article{maddison2016concrete,
  title={The concrete distribution: A continuous relaxation of discrete random variables},
  author={Maddison, Chris J and Mnih, Andriy and Teh, Yee Whye},
  journal={arXiv preprint arXiv:1611.00712},
  year={2016}
}


% George Kour and Raid Saabne.
% \newblock Real-time segmentation of on-line handwritten arabic script.
% \newblock In {\em Frontiers in Handwriting Recognition (ICFHR), 2014 14th
%   International Conference on}, pages 417--422. IEEE, 2014.

% George Kour and Raid Saabne.
% \newblock Fast classification of handwritten on-line arabic characters.
% \newblock In {\em Soft Computing and Pattern Recognition (SoCPaR), 2014 6th
%   International Conference of}, pages 312--318. IEEE, 2014.

% \bibitem{hadash2018estimate}
% Guy Hadash, Einat Kermany, Boaz Carmeli, Ofer Lavi, George Kour, and Alon
% \newblock Estimate and replace: A novel approach to integrating deep neural
%   networks with existing applications.
% \newblock {\em arXiv preprint arXiv:1804.09028}, 2018.

% \end{thebibliography}

\end{document}